\documentclass{article}

\usepackage{microtype}
\usepackage{graphicx}
\usepackage{subfigure}
\usepackage{subcaption}
\usepackage{tcolorbox}

\usepackage{booktabs} 

\usepackage{hyperref}

\newcommand{\llamaeight}{Llama-3.1-8b}
\newcommand{\llamaseventy}{Llama-3.1-70b}
\newcommand{\mixtral}{Mixtral-8x22B}
\newcommand{\claude}{Claude-3.5-Sonnet}
\newcommand{\deepseek}{DeepSeek-V3}
\newcommand{\gpt}{GPT-4o}

\usepackage[accepted]{icml2025_arxiv}

\usepackage{amsmath}
\usepackage{amssymb}
\usepackage{mathtools}
\usepackage{amsthm}
\usepackage{epigraph}
\newcommand{\xhdr}[1]{\vspace{1.7mm}\noindent{{\bf #1.}}}

\usepackage[capitalize,noabbrev]{cleveref}

\theoremstyle{plain}

\theoremstyle{definition}

\theoremstyle{remark}

\usepackage[textsize=tiny]{todonotes}

\icmltitlerunning{What is a Number, That a Large Language Model May Know It?}

\begin{document}

\twocolumn[
\icmltitle{What is a Number, That a Large Language Model May Know It?}



\icmlsetsymbol{equal}{*}

\begin{icmlauthorlist}
\icmlauthor{Raja Marjieh}{equal,psych}
\icmlauthor{Veniamin Veselovsky}{equal,cs}
\icmlauthor{Thomas L. Griffiths$^\dagger$}{psych,cs}
\icmlauthor{Ilia Sucholutsky$^\dagger$}{nyu}
\end{icmlauthorlist}

\icmlaffiliation{psych}{Department of Psychology, Princeton University}
\icmlaffiliation{cs}{Department of Computer Science, Princeton University}
\icmlaffiliation{nyu}{Center for Data Science, New York University}

\icmlcorrespondingauthor{Raja Marjieh}{raja.marjieh@princeton.edu}

\icmlkeywords{Numbers, Representations, Cognitive Science, LLMs}

\vskip 0.3in
]



\printAffiliationsAndNotice{$^\dagger$\icmlEqualContribution} 


\begin{abstract}
Numbers are a basic part of how humans represent and describe the world around them. As a consequence, learning effective representations of numbers is critical for the success of large language models as they become more integrated into everyday decisions. However, these models face a challenge: depending on context, the same sequence of digit tokens, e.g., 911, can be treated as a number or as a string. What kind of representations arise from this duality, and what are its downstream implications? Using a similarity-based prompting technique from cognitive science, we show that LLMs learn representational spaces that blend string-like and numerical representations. In particular, we show that elicited similarity judgments from these models over integer pairs can be captured by a combination of Levenshtein edit distance and numerical Log-Linear distance, suggesting an entangled representation. In a series of experiments we show how this entanglement is reflected in the latent embeddings, how it can be reduced but not entirely eliminated by context, and how it can propagate into a realistic decision scenario. These results shed light on a representational tension in transformer models that must learn what a number is from text input.
\end{abstract}

\section{Introduction}
\epigraph{``What is a number, that a man may know it, and a man, that he may know a number?''}{\textit{Warren McCulloch} (\citeyear{mcculloch1961number})}

Numbers play a pivotal role in many aspects of human cognition \cite{dehaene2011number}. Starting from a narrow sense of magnitude in infancy, humans gradually acquire nuanced representations of number that capture abstract properties such as ``even'' and ``odd'' \cite{miller1983child,piantadosi2014children}. Decades of research have been devoted to understanding how humans process and represent numbers \cite{miller1983child,nieder2009representation,cheyette2020unified,tenenbaum1999rules,piantadosi2016rational}. In fact, this endeavor dates back to some of the earliest applications of neural networks to artificial intelligence \cite{mcculloch1943logical,mcculloch1961number} when researchers sought to define computing machines (a simulated ``man'') that can intelligently process numbers. 

The question of how computing machines represent number assumes key importance as modern neural networks, in particular large language models (LLMs), are becoming integrated into everyday decisions \cite{zhu-etal-2025-language,hanna2023how,stolfo2023mechanisticinterpretationarithmeticreasoning,mccoy2023embersautoregressionunderstandinglarge,mccoy2024languagemodeloptimizedreasoning,tian2023just}. Since these models learn representations by predicting tokens in text, LLMs face a challenge: depending on context, the same sequence of digit tokens, e.g., 101 or 911, can be treated as a number or as a string. This duality introduces an issue akin to polysemy and homonymy \cite{vicente2017polysemy}, but it extends across different symbolic systems. What kind of representations arise from this duality, and what are its downstream implications? This is not straightforward to answer, in part because LLMs lack the experience that shapes humans' number representations, and in part because model internals are not always available and diagnostic tests can be challenging to design.

Here we address this challenge by leveraging tools from cognitive science for characterizing representations \cite{shepard1980multidimensional,shepard1987toward,tversky1986nearest,tenenbaum2001generalization,marjieh2024universal}. Specifically, we elicit similarity judgments across number pairs from six modern LLMs using an appropriate prompt. We then use those judgments to construct detailed maps (similarity matrices) that capture how the models organize numbers. Crucially, this technique can applied to any model without the need to access its internals. Moreover, when the model internals are available, this approach can be combined with probing techniques to evaluate how prompt behavior is reflected in the structure of the latent embeddings.

\begin{figure*}[ht]
    \centering
    \includegraphics[width=0.9\linewidth]{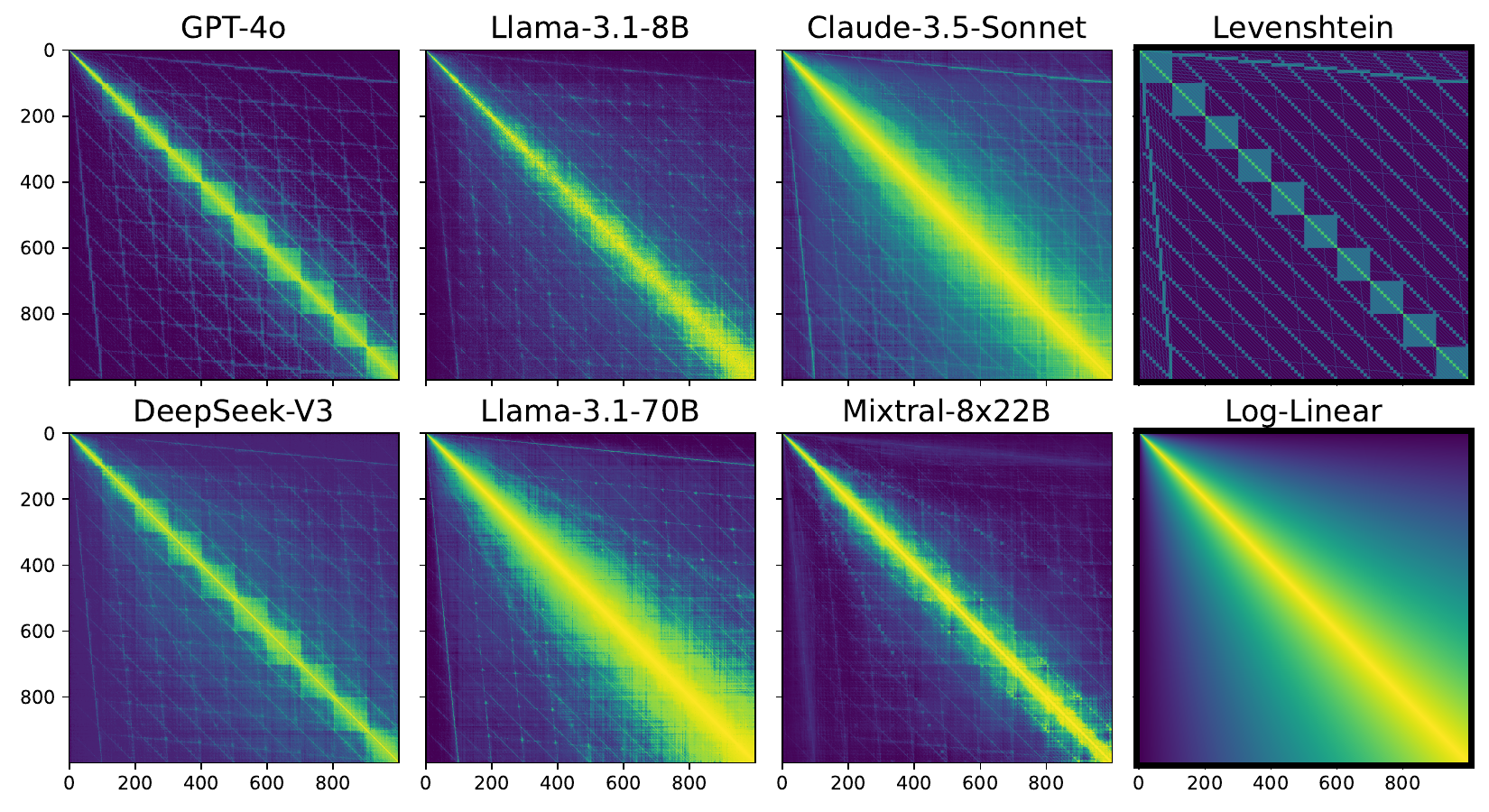}
    \caption{LLM number similarity matrices (symmetrized) over all integer pairs in the range $0-999$, along with two theoretical similarity matrices derived from a Levenshtein string edit distance and a psychological Log-Linear numerical distance (highlighted in black).}
    \label{fig:default-sim}
\end{figure*}

We show that despite variation in model training data and size, all models exhibit highly regular patterns that can be adequately decomposed in terms of a psychologically-motivated numerical distance based on a Log-Linear representation of numbers \cite{piantadosi2016rational} and a string edit distance, suggesting an entanglement. We show how this entanglement can be reduced but not eliminated by introducing a context that specifies the `type' of the integer (\texttt{int()} vs. \texttt{str()}), and how it can be probed internally in a model's latent space. Inspired by these results we then construct a decision scenario that reveals how string-bias can propagate into a realistic setting leading to incorrect answers. Viewed together, these findings shed light on an intrinsic tension between number and string representations that modern large language models must learn to navigate.

\section{Related Work}
\subsection{Numbers and Language Models}
The processing and representation of numbers in large language models is a topic of active research. Number representations have recently been studied within pretrained language models using techniques from mechanistic interpretability. For example, \citet{zhu-etal-2025-language} devised a dataset of addition problems to show that LLMs encode the value of numbers linearly in that context. Likewise, circuit analysis has been adopted to characterize how smaller language models compute ``greater than''~\cite{hanna2023how} and basic arithmetic operations \cite{stolfo2023mechanisticinterpretationarithmeticreasoning}. Other work studied how simple models solve modular arithmetic tasks~\cite{nanda2023progressmeasuresgrokkingmechanistic}.  
More behavioral approaches have also been explored. \citet{tian2023just} looked at how model generations of numbers can be used to test for answer calibration. Likewise, \citeauthor{mccoy2023embersautoregressionunderstandinglarge} (\citeyear{mccoy2023embersautoregressionunderstandinglarge,mccoy2024languagemodeloptimizedreasoning}) showed how a Bayesian framework can be used to characterize how LLM performance on various tasks, including numeric ones, depends on the probability of the input, task, and output.

\subsection{Probing and Representation Analysis in LLMs}
Probing is a well-established technique for studying the internal representations of machine learning models~\cite{alain2018understandingintermediatelayersusing,belinkov2021probingclassifierspromisesshortcomings}. Dating back to BERT~\cite{devlin2019bertpretrainingdeepbidirectional}, probing has been used to explore various aspects of representation such as how language models encode sentence structure~\cite{tenney2019you,manning2020emergent,bertology}. More recently, probing has become a common technique in studying larger language models. In the work by \citet{zhu-etal-2025-language} mentioned earlier, the authors designed linear probes to extract number value from latent embeddings. Probes have also been used to discover how language models process concepts across layers~\cite{gurnee2023finding}, and for extracting uncertainty in language model generations~\cite{kadavath2022language}. 
Similarly, probes, and latent analysis more broadly, have been used in the world models literature, e.g., to show that language models trained on Othello games develop a world model of the game~\cite{nanda-etal-2023-emergent}, and that they learn to represent space and time~\cite{gurnee2024languagemodelsrepresentspace,gurnee2023finding,nylund2023time}.
More recent approaches to studying representations in language models have also been proposed like activation patching~\cite{meng2022locating}, sparse autoencoders~\cite{cunningham2023sparse,gao2024scaling}, and distributed alignment search~\cite{geiger2024findingalignmentsinterpretablecausal}. 






\subsection{Behavioral Analysis of Language Models}
By combining the prompt comprehension capabilities of large language models with the wide range of paradigms available in the behavioral sciences, a growing line of work proposes new tools for characterizing behavior in LLMs. For example, \citet{marjieh2024large} used a suite of perceptual judgment tasks from psychophysics to characterize sensory knowledge in LLMs. \citet{binz2023using} leveraged paradigms from cognitive psychology to study how GPT-3 solves various tasks. \citet{bai2024measuring} used tools from social psychology to diagnose implicit racial and gender bias in LLMs. \citet{webb2023emergent} subjected LLMs to analogical tasks to study their abstract reasoning capacities. Our work takes a similar approach to this line of work, and combines it with modern probing techniques to study LLM behavior and representation in the domain of number.

\begin{figure*}[ht]
    \centering
    \includegraphics[width=0.85\linewidth]{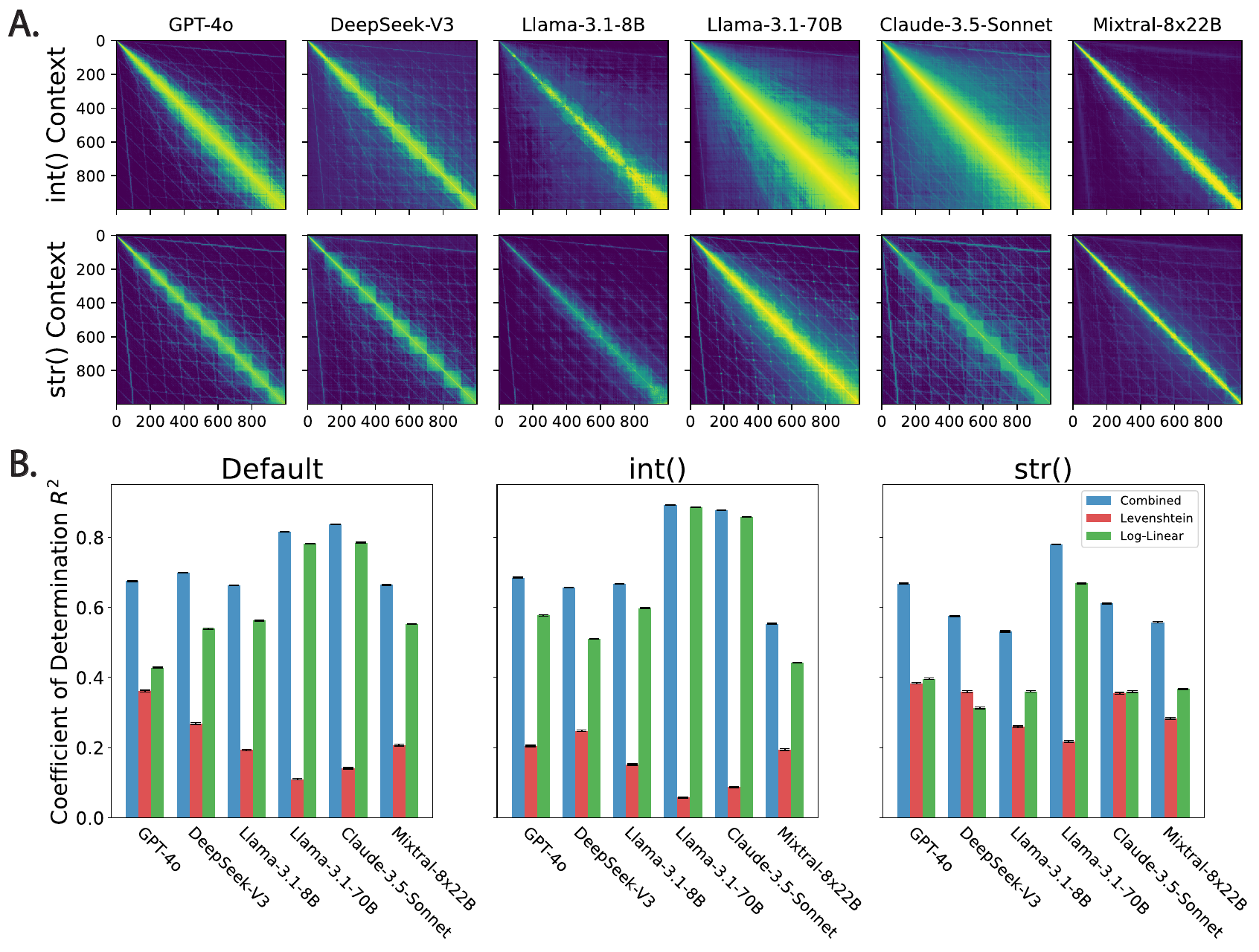}
    \caption{Context effects on LLM-elicited number similarity matrices and their decomposition. \textbf{A}. LLM similarity matrices under the effect of `type' specification: \texttt{int()} vs. \texttt{str()} (see Appendix \ref{app:prompts} for prompts). \textbf{B}. Coefficient of determination ($R^2$) for the different similarity matrices under the default (Figure~\ref{fig:default-sim}), \texttt{int()}, and \texttt{str()} contexts for the combined and separate Levenshtein (string) and Log-Linear (numerical) distance predictors (error bars indicate 95\% confidence intervals; see Methodology).}
    \label{fig:int-str-sim}
\end{figure*}

\section{Probing Number Representations in LLMs with Similarity Judgments}

Our approach builds on the paradigm of similarity judgments from cognitive science \cite{shepard1962analysis,shepard1980multidimensional,tversky1986nearest,tenenbaum2001generalization,marjieh2024universal}, which is also closely related to representational similarity analysis (RSA) from neuroscience \cite{kriegeskorte2008representational}. Given a domain of interest $\mathcal{D}$, a set of representative items from that domain $\{x_1,x_2,\dots,x_N\}\subset\mathcal{D}$ (or `stimuli'), and an agent $\mathcal{A}$ whose representation $\mathcal{M}(\mathcal{D})$ one would like to characterize, the paradigm proceeds by eliciting pairwise similarity judgments from $\mathcal{A}$ across all pairs of items (`How similar is the item $x_i$ to the item $x_j$?'). These judgments are then aggregated into a similarity matrix $s_{ij}$ which defines a notion of proximity on $\mathcal{D}$ that can be used to characterize $\mathcal{M}(\mathcal{D})$. Since the notion of similarity is by construction neutral, it leaves it to the agent to impose its own structure on $s_{ij}$.

In our case, the domain of interest is the set of non-negative integers $\{0,1,2,\dots \}$, and the agent is the LLM in question. We consider a representative sample of state-of-the-art models, namely, \gpt{} \cite{hurst2024gpt}, two variants of Llama-3.1 (8b and 70b) \cite{dubey2024llama}, \deepseek{} \cite{liu2024deepseek}, \claude{} \cite{anthropic2024claude}, and \mixtral{} \cite{jiang2024mixtral}. We then apply the similarity technique in a series of experiments that span different contexts as well as behavioral (prompt-level) and internal (embedding-level) analyses and examine how the observed patterns decompose in terms of theoretical string and numerical metrics. We detail those experiments in what follows.

\section{Methodology}
\subsection{Tasks}
Our experiments fall into three categories, all of which are associated with a certain behavioral prompt for eliciting LLM judgments over numerical quantities. The prompts are provided in Appendix \ref{app:prompts}. In the first category, we elicited similarity judgments over all pairs of numbers in the range $0-999$  (`How similar are the two numbers on a scale of 0 (completely dissimilar) to 1 (completely similar)?'; see Appendix \ref{app:prompts}). We then constructed symmetric similarity matrices $s_{ij}$ by averaging over both presentation orders (i.e., $s(x_i,x_j)$ and $s(x_j,x_i)$), and then evaluated the structure that emerged (see Evaluation Metrics below) and the way it is affected by context. We focused on the range $0-999$ because all integers within it are represented as unique tokens in the models considered, which controls for any tokenization-specific differences. In the second category, we extended the similarity analysis for one of the models for which we had internal access (\llamaeight) and probed the extent to which its internal token embeddings reflected the external behavior through a linear transformation (see Probing Language Models). Finally, in the third category we constructed a naturalistic decision scenario in which the model had to select from two available numerical quantities $q_1,q_2$ (a compound concentration in a test tube) the one that is most similar to a desired quantity $q_0$ (further details in Experiments and Appendix \ref{app:prompts}). We constructed such triplets $(q_0,q_1,q_2)$ in a way that the answer would diverge depending on whether the quantities are treated as numbers or strings (see Constructing Close Triplets below) which then allowed us to probe the string-bias of models.

\subsection{Models}
We evaluated four open-source and two closed-source models. The open source models consist of \llamaeight{}-Instruct, \llamaseventy{}-Instruct, \deepseek{}, \mixtral{}-Instruct-V0.1, and the closed source include \gpt{} and \claude{}. Henceforth, we will refer to the Llama and Mixtral models as \llamaeight{}, \llamaseventy{}, and \mixtral{} to improve readability. 
Since generating $1000\times1000$ similarities requires sampling 1,000,000 pairwise comparisons which costs around \$160 for Claude-3.5-Sonnet, we limit the model evaluations across all contexts to one run at a temperature of zero to get the most likely answer across the different settings. For comparison, we include results from an additional run in the basic similarity context at a temperature of 0.7 in Appendix~\ref{app:extensions}.


\subsection{Evaluation Metrics}
To characterize the extent to which the derived similarity matrices $s_{ij}$ were string-like or number-like, we regressed their values against two theoretical distance measures, namely, the Levenshtein string edit-distance $d_{Lev}(a,b)$ \cite{levenshtein1966binary} and the psychological Log-Linear number distance $d_{Log}(a,b)$ \cite{piantadosi2016rational} (we also considered a simple linear $\ell_1$ distance $|a-b|$ but found that it led to worse results; Appendix Figure \ref{fig:lin-alt}). The Log-Linear distance captures the psychological phenomenon whereby humans represent larger numbers with less fidelity leading to a logarithmic sensitivity to magnitude, i.e. $x-y \rightarrow \log(x)-\log(y)$ (see Appendix \ref{app:lev} for definition). The Levenshtein distance, on the other hand, is defined as the minimum number of deletions, insertions, or substitutions required to transform one string to the other. This can be defined recursively (see Appendix \ref{app:lev}) and we used the Python package \texttt{Levenshtein}\footnote{\url{https://rapidfuzz.github.io/Levenshtein/}} to compute it. For example, the Levenshtein distance between the digits $200$ and $100$ is 1 because it takes a single substitution ($2\rightarrow 1$) to transform one string into the other. 

Given the two distance measures evaluated on the range of interest $d_{ij}^{(Lev)}$ and $d_{ij}^{(Log)}$, we transformed them into similarity scores $s_{ij}(d)=1-d_{ij}/\max\{d_{ij}\}$ and then fitted them to $s_{ij}$ (after z-scoring), i.e., $s_{ij}=\alpha + \beta s_{ij}^{(Lev)} + \gamma s_{ij}^{(Log)}$ using the LinearRegression method from \texttt{scikit-learn} \cite{scikit-learn} and computed the coefficient of determination ($R^2$ or variance-explained) which we also repeated for the individual metrics separately. To derive confidence intervals, we bootstrapped over the number pairs (since the models had zero temperature) with replacement and 1,000 repetitions. All analysis code is provided in the accompanying repository (see Reproducibility).

\subsection{Probing Language Models}
When generating an output, a language model first tokenizes the input sentence into a sequence of tokens $x_1,...,x_n\in V$ where $V$ is some vocabulary. Then during the forward pass, the model incrementally transforms the initial embeddings (extracted from a vocabulary embedding layer) through each layer. 
The internal representations at each layer --- commonly referred to as \emph{residuals} --- capture evolving latent features. These residuals can be analyzed to probe specific properties of the input or model behavior.

In our case, we train an affine transformation for each layer within the model that takes in a specific latent $h_i^{(j)}$ for token $i$ at layer $j$, and then learns a vector $w$ and bias term $b$, such that $w \cdot h_i^{(j)} + b$  predicts a specific distance measure. 
Due to computational constraints, the probes are run only on the \llamaeight{} model. We take the original similarity prompt and extract the residuals from the last token --- in our case the ``:'' after ``Result:'' (see Appendix \ref{app:prompts}) --- and train two linear probes on them. One probe is trained to predict the Levenshtein distance between the two inputted numbers, the other the Log-Linear distance.

We train the probes on 9,500 random pairs of numbers between 0 and 999, and evaluate on the range 0-500. We ablate across all layers and select layer 25 for the final probe (see Appendix~\ref{app:probing} for ablations on number of training points and performance across layers). We then run these probes on all 500$\times$500 (250,000) combinations of number similarities and report their corresponding correlations. To illustrate the results of the probe, we apply multidimensional scaling (MDS) \cite{shepard1962analysis} on the 500$\times$500 similarity judgments extracted by the probe. MDS is a visualization technique that represents high-dimensional data in a lower-dimensional space, preserving the relative pairwise distances or dissimilarities as closely as possible. We computed MDS embeddings using \texttt{scikit-learn}.


\begin{figure*}[ht]
    \centering
    \includegraphics[width=0.65\linewidth]{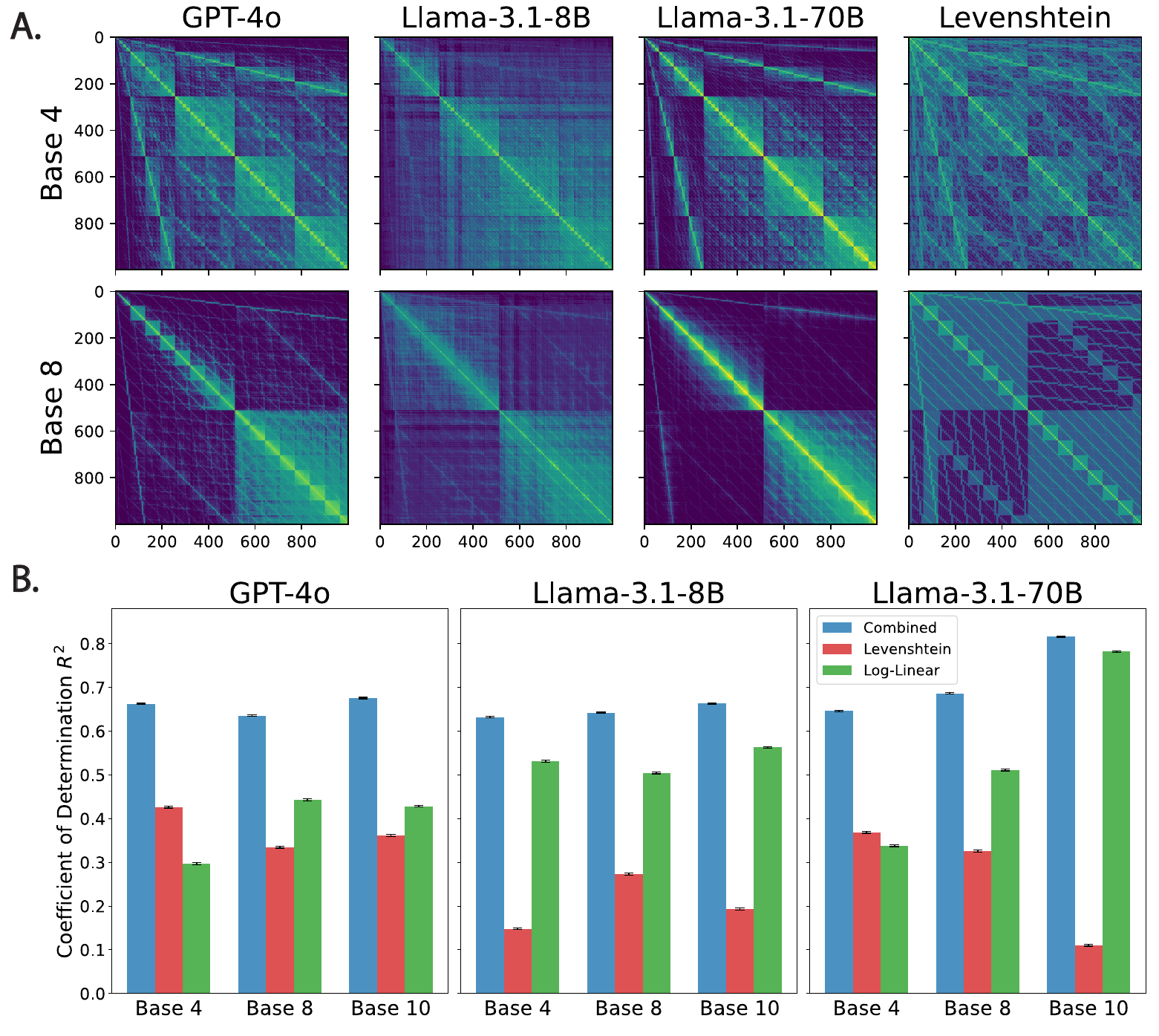}
    \caption{The effect of other number bases on elicited similarity. \textbf{A.} LLM similarity matrices over all integer pairs in the range $0-999$ represented in base 4 and 8 along with the corresponding Levenshtein distance measures (see Appendix \ref{app:prompts} for prompts). \textbf{B.} Coefficient of determination ($R^2$) for the various similarity matrices under the different base contexts (including the base 10 results from Figure~\ref{fig:default-sim}) for the combined and separate Levenshtein (string) and Log-Linear (numerical) distance predictors (error bars indicate 95\% CIs).}
    \label{fig:other-bases}
\end{figure*}
\subsection{Constructing Close Triplets}
Our goal was to construct diagnostic triplets for the naturalistic decision scenario such that (i) the options are numerically close to each other but there is an unambiguous correct answer, and (ii) the options are very different in terms of their edit distance relative to the target. To do that, we constructed the target quantity $q_0$ by randomly sampling three digits from the set $\{2,\dots,9\}$ with replacement and combining them into a number (e.g., $3,3,1\rightarrow331$). Then, we constructed a Levenshtein-aligned option by subtracting 1 from the largest decimal entry (i.e., $331\rightarrow231$). Finally, we constructed a Log-aligned option by keeping the largest decimal entry from the target and randomly sampling two new integers from the range $\{1,\dots,9\}$ excluding the digits that appeared in the other two numbers (e.g., $357$). Thus, the result would be $(331,231,357)$ in our example which satisfies the desired properties. We repeated the process also for five digit numbers to probe how things change for longer numbers (e.g., $25337,15337,26886$). We sampled 10,000 such triplets from each length category and took the unique subset. Overall, there were 6,474 unique three digit triplets, and 9,995 five digit triplets. 

\section{Experiments}
\begin{figure*}[ht]
    \centering
    \includegraphics[width=.75\linewidth]{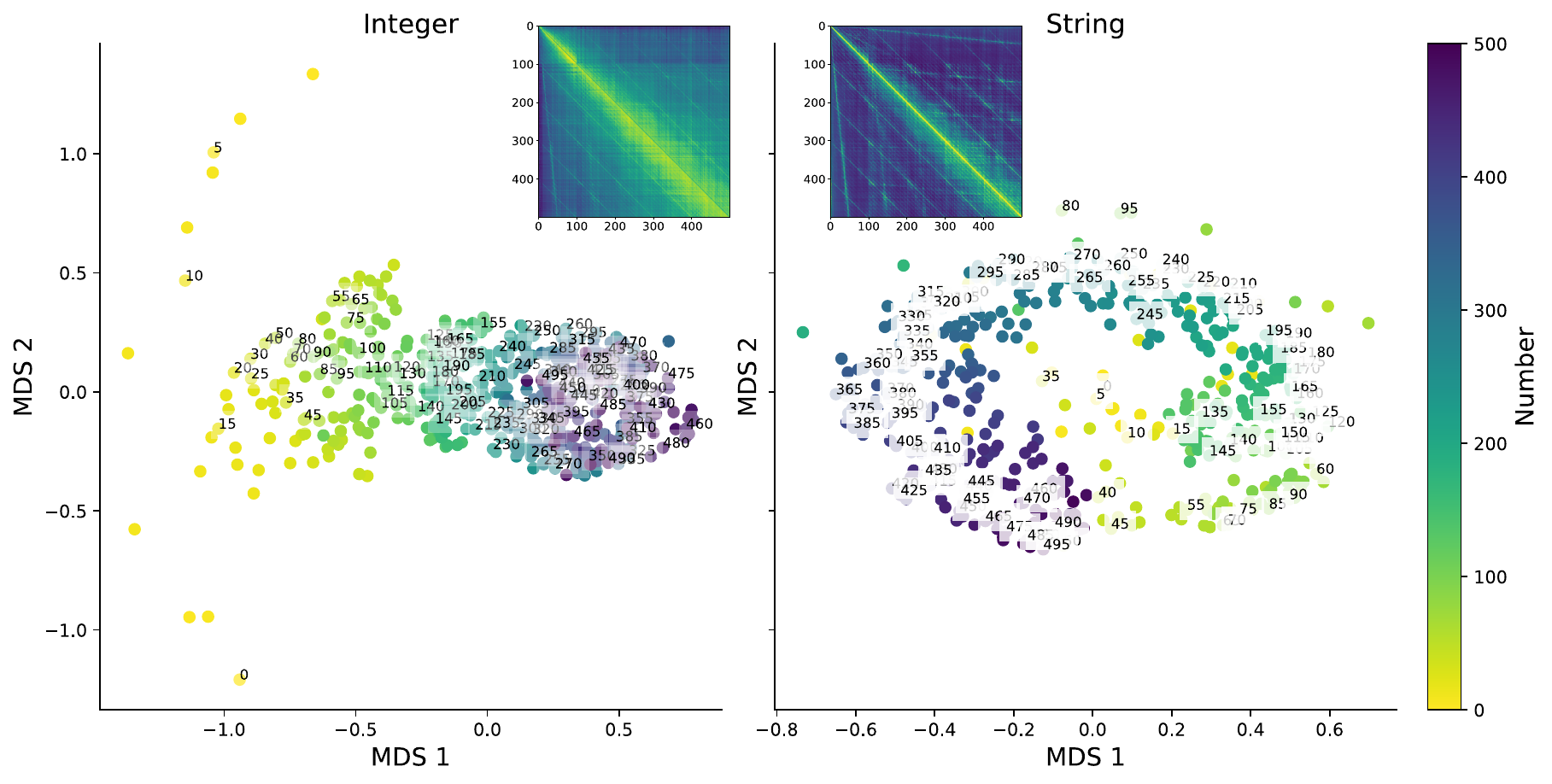}
    \caption{Decoded string and integer subspaces from \llamaeight{} using linear probes (see Methodology). The decoded similarity matrices are provided as insets along with their multidimensional scaling solutions (MDS). Integers are labeled every 5 points.}
    \label{fig:mds}
\end{figure*}
\subsection{Eliciting Similarity Judgments}
In the first experiment, we elicited similarity judgments across all pairs of integers in the range $0-999$ (`How similar are the two numbers?'; see Appendix \ref{app:prompts} for the full prompt). We begin by presenting the qualitative patterns in the raw data and then quantify them using a suitable regression analysis. Figure~\ref{fig:default-sim} shows the similarity matrices for all models (see Methodology). The resulting LLM matrices are highly structured with a dominant area around the diagonal that in some cases (e.g., \gpt{} and \deepseek) takes a block-diagonal form, in addition to other sub-diagonal structures. These patterns appear to blend the properties found in two theoretical similarity matrices associated with string edit distance (Levenshtein) and numerical  distance (Log-Linear) highlighted in the rightmost panels of Figure~\ref{fig:default-sim} (see Methodology). To further see whether these patterns arise from a tension between string and integer representations, in a following experiment we resolved the ambiguity of the similarity prompt by specifying the `type' of the token using \texttt{int()} and \texttt{str()} contexts (see Appendix~\ref{app:prompts} for prompts). The results are shown in Figure~\ref{fig:int-str-sim}A. In this case, we found that the context intervention pushed the matrices in opposing directions. For example, \gpt{} loses its block diagonal structure in the \texttt{int()} context, whereas \claude{} gains it in the \texttt{str()} context. Similar trends can also be found in the other models.

To quantify the above, we regressed the Levenshtein and Log-Linear distance measures against the LLM similarity judgments in the  three contexts considered. A breakdown of the explained variance (coefficient of determination $R^2$) per condition are provided in Figure~\ref{fig:int-str-sim}B. Overall, in the default context we found that a linear combination of the numerical and Levenshtein distance measures is sufficient to achieve an average $R^2$ of $.726$ (95\% CI of mean: $[.725,.726]$), with the separate components each contributing significantly (Log-Linear only $R^2$ CI: $[.607,.609]$, Levenshtein only $R^2$ CI: $[.213,.215]$). We note that replacing the Log-Linear distance with a simple linear $\ell_1$ distance diminishes combined average performance to an $R^2$ of $.567$ (95\% CI: $[.567,.568]$; see Appendix Figure~\ref{fig:lin-alt}). In the other contexts, the metrics were correspondingly pushed in opposing directions. In the \texttt{str()} case, the mean $R^2$ CIs were: Combined: $[.620,.621]$, Log-Linear: $[.410,.412]$, and Levenshtein: $[.309,.311]$. On the other hand, in the \texttt{int()} context these were: Combined: $[.721,.722]$, Log-Linear: $[.645,.646]$, and Levenshtein: $[.156,.158]$. 


Next, we evaluated the extent to which the string-like similarity generalizes to less common number bases. The idea here is that if the model is relying on edit distance, then the effect should persist irrespective of how uncommon the underlying numerical basis is. To that end, we elicited additional similarity judgments for a subset of three models (\gpt{}, \llamaeight{}, and \llamaseventy{}) using two additional number bases: base 4 and base 8 (see Methodology). The results are shown in Figure~\ref{fig:other-bases}A. Here too we found highly structured matrices with patterns that are consistent with those derived from the Levenshtein distance. Quantitatively, we found that all bases resulted in a significant Levenshtein contribution (Figure \ref{fig:other-bases}B), and in the case of \gpt{} and \llamaseventy{} those contributions were enhanced relative to the Log-Linear metric. Indeed, compare the $R^2$ CIs for the Log-Linear vs. Levenshtein regressors: In the case of \llamaseventy{}, for base 10 we had: Log-Linear: $[.780,.783]$ and Levenshtein: $[.108,.112]$, whereas for base 4 this is reversed: Log-Linear: $[.336,.340]$, and Levenshtein: $[.366,.371]$. Likewise, for \gpt{} in base 10: Log-Linear: $[.426,.430]$, and Levenshtein: $[.359,.364]$, whereas for base 4: Log-Linear: $[.295,.299]$, and Levenshtein: $[.424,.428]$.

\subsection{Probing Internal Representations}
The previous experiments showed that when an LLM is prompted for similarity judgments between two different integer tokens, the resulting behavioral metric is a combination of string and integer distance. In this section, we show how this also holds on the representational (i.e., token-embedding) level in a model for which we have internal access (\llamaeight{}). To do that, we train linear probes from the last token in the prompt to decode the Log-Linear and Levenshtein distances between the integers in question (see Methodology). This is non-trivial as there is no guarantee that such a linear transformation exists unless the model encodes the relevant information in its embedding (this can be seen also from the fact that our probes had 4,096 parameters, corresponding to the latent dimension, fitted on 9,500 data points, and then evaluated on 250,000 data points; see Methodology). In Figure~\ref{fig:mds} we illustrate the decoded similarity matrices from the embeddings. Here we notice a pattern that is consistent with the behavioral (prompt-based) data. In the string probe case (right panel in Figure~\ref{fig:mds}), the model is able to capture the edit-distance similarity between two numbers; whereas in the integer probe case (left panel in Figure~\ref{fig:mds}), the decoded pattern is indeed much more log-linear as can be seen from the diagonal area in the matrix. Interestingly, however, string similarities still bleed into the representation as can be seen from the sub-diagonal structures. Quantitatively, the Pearson correlation between the string probe and the Levenshtein and Log-Linear measures is .650 and .527, respectively. For the integer probe, the correlation is .917 with the Log-Linear distance and .393 with the Levenshtein distance (c.f., Table \ref{tab:correlations} in Appendix~\ref{app:probing}).

To further highlight the structure of the latent subspaces, in Figure~\ref{fig:mds} we apply multidimensional scaling \cite{shepard1980multidimensional} to the decoded similarity matrices in Figure~\ref{fig:mds} to derive two-dimensional maps of the relations between integers. We see that the probes nicely dissociate the integer and string subspaces. In the integer subspace, numbers are organized on a scale that captures numerical distance. On the other hand, in the string subspace, the scale is replaced with a non-linear pattern that closely follows edit distance. 


\subsection{Behavioral Implications}
In the final experiment we wanted to see whether there are behavioral implications for the number-string tension in a naturalistic scenario. To do that, we presented the models with an empirical setting in which they required a test tube with a certain unavailable compound concentration (given in parts per million units; ppm). The models were asked to choose from two available test tubes the one containing the most similar concentration to the one desired. Crucially, we constructed diagnostic concentration triplets that lead to divergent answers depending on whether a Levenshtein distance is used or a Log-Linear one (see Methodology). Here is an example prompt: `You require a compound with a concentration of approximately 785 ppm. Two test tubes are available: one containing 685 ppm and the other 791 ppm. Your task is to determine which test tube provides the most similar concentration to your required dosage.' Clearly, a concentration of 791 ppm is much more similar to 785 ppm than 685 ppm, but a model with strong string bias may erroneously choose 685 ppm which shares more digits with 785 ppm. We then evaluated the percentage of incorrect (Levenshtein-consistent) answers as a measure of string bias. In addition to the 3-digit triplets, we also considered 5-digit ones to see whether the problem may be exacerbated by the inclusion of more digits (e.g., 22565, 12565, and 28743; see Methodology). We also considered both orders of presentation to see whether there are any ordering effects when the Levenshtein-aligned option is presented first vs. the Log-Linear one (`Reverse' setting). 
\begin{figure*}[!ht]
    \centering
    \includegraphics[width=0.7\linewidth]{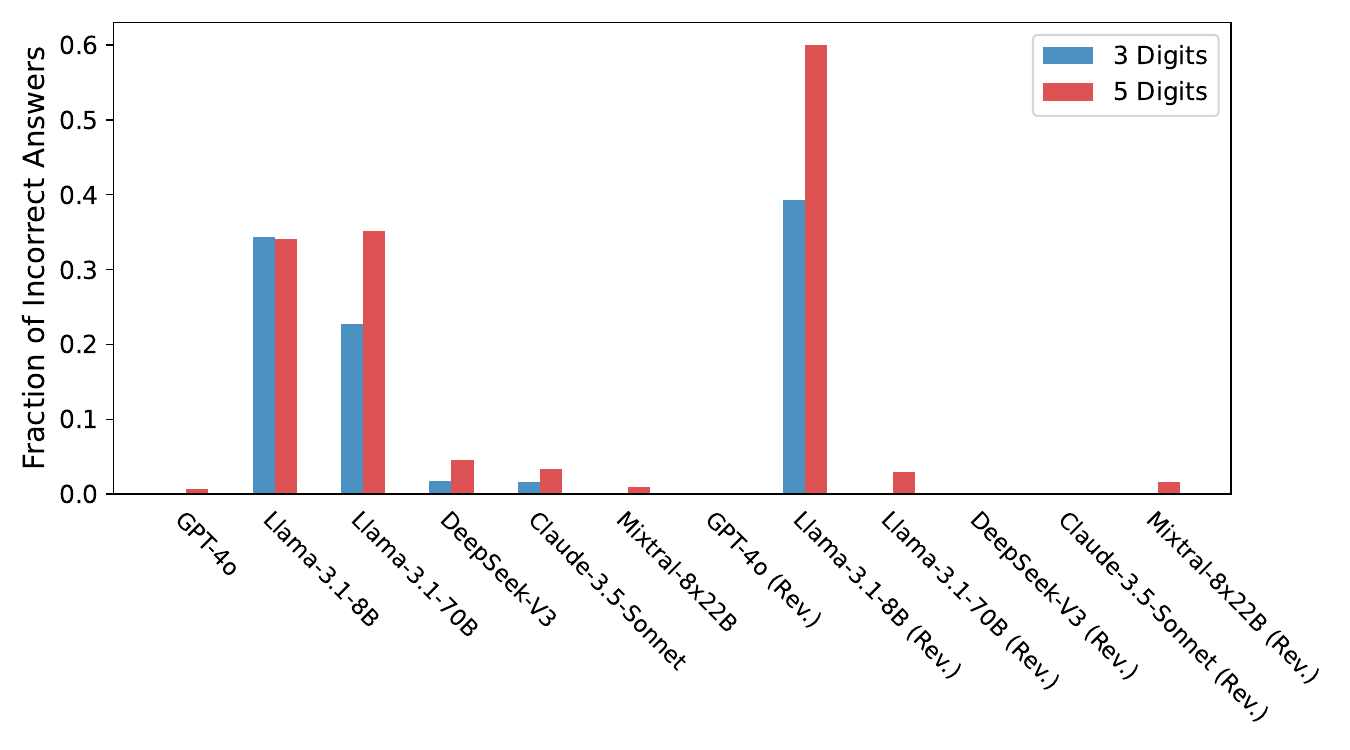}
    \caption{Probing string-bias in a naturalistic decision scenario. Bar plots indicate the faction of times an incorrect (Levenshtein-aligned) option was chosen for the 3-digit and 5-digit scenarios considered, and the two possible presentation orders (see Methodology). `(Rev.)' indicates the case in which the Levenshtein-aligned option was presented second (see prompt in Appendix~\ref{app:prompts}).}
    \label{fig:realistic-context}
\end{figure*}
The results are shown in Figure~\ref{fig:realistic-context}. We found that all models had some degree of string-bias errors but the range varied greatly and was enhanced for longer numbers. In the 3-digit case the largest percentage (averaged over order) was 36.86\% for \llamaeight{} (followed by 11.38\% for \llamaseventy{}), and the smallest was 0.02\% for \gpt{}. Likewise, in the 5-digit case, the largest percentage was 47.03\% for \llamaeight{} (followed by 19.04\% for \llamaseventy{}), and the smallest was 0.41\% for \gpt{}. Interestingly, the Llama-3.1 models exhibited substantially larger degrees of bias, as well as some degree of order effects in some cases (note that this cannot be attributed to a generic presentation bias since in that case we wouldn't expect to see a difference between the 3-digit and 5-digit scenarios). Overall, these results suggest that the number-string tension can manifest itself in a realistic context, and that the extent that such a context can suppress the stringiness behavior varies across models.


\section{Discussion}
We have provided evidence for a representational tension in large language models that learn to represent numbers by processing text input. This tension is reflected externally (through prompt-elicited similarity judgments), internally (through embedding probes), and could also propagate into a quantitative decision scenario.

Our results suggest that the tension may arise from ambiguity in the `type' of digit tokens in a given context. Specifically, by explicitly specifying the type of the tokens in the prompt (i.e., \texttt{int()} vs. \texttt{str()}) we showed how the similarity matrices in the default case can be pushed to align more strictly with string or log-linear distance. This ambiguity is reminiscent of polysemy and homonymy in language \cite{vicente2017polysemy} whereby a single word can have multiple meanings, but in our case the ambiguity is extended across different systems. This entanglement of types is also reflected in our probing results whereby the linearly decoded similarity matrices had some residual shared structure.

Our findings are consistent with recent probing work showing that LLMs are able to encode the value of numbers \cite{zhu-etal-2025-language}. Interestingly, however, we found that when the setting is not strictly mathematical as in arithmetic, LLMs appear to exhibit a more psychologically plausible log-linear representation, though blended with string distance. This is consistent with recent work showing that LLMs encode rich sensory knowledge \cite{marjieh2024large}. The results also make contact with the literature on how LLMs perform numerical operations \cite{hanna2023how,stolfo2023mechanisticinterpretationarithmeticreasoning} and the possible sources of numerical errors that could arise from their mode of learning \cite{mccoy2023embersautoregressionunderstandinglarge,mccoy2024languagemodeloptimizedreasoning}. One possibility is that LLMs attempt to categorize the type of an integer token from context, and then use that to specify a set of operations on units of that type. For example, if an LLM views the numbers $1$ and $2$ as strings, then asking it to sum the two numbers may result in $12$ rather than $3$. Future work could interrogate whether there is indeed a causal link between such type-categorization mechanisms and numerical errors.

We end by discussing some limitations that point towards future directions. First, due to resource limitations our LLM results were restricted to zero temperature (though see Appendix~\ref{app:extensions}) and the probing analysis was limited to the smaller \llamaeight{} model. Future work could look into how the results generalize to a stochastic sampling setting and other models for probing. Second, while we explored the effect of context on number similarity judgments, it remains to be seen whether one could causally intervene at the embedding level to steer the model from one token type to another. Third, while our naturalistic decision scenario was designed to be as diagnostic as possible, further work is necessary to assess how pervasive this issue is in more generic daily numerical use cases. 
Relatedly, reasoning-based models are increasingly becoming the \emph{de facto} standard in user-facing applications (e.g., OpenAI’s o1~\citealp{jaech2024openai}, DeepSeek’s R1~\citealp{deepseekai2025deepseekr1incentivizingreasoningcapability}). Extending our findings to these models is a promising avenue for future work, as their ability to reason may fundamentally alter how similarity judgments are produced.
Finally, our work was restricted to the domain of numbers but our results may be a special case of a more broad symbol-string duality that impacts other domains (e.g., code and special characters). We hope to report on these directions in the future. 


As artificial intelligence becomes increasingly intermingled with human intelligence, knowing what a number is becomes an increasingly important concern. \citet{mcculloch1961number} argued that the human capacity to understand numbers could be acquired by machines based on artificial neural networks. Our results suggest that there are still meaningful differences in how humans and large language models based on artificial neural networks conceive of numbers.

\xhdr{Reproducibility} Code for reproducing the analyses can be accessed here: \url{https://github.com/vminvsky/numbers-in-llms}

\xhdr{Acknowledgments} This work was supported by an Azure Foundation Models Research grant from Microsoft.


\label{submission}

\bibliography{main}
\bibliographystyle{icml2025}

\newpage
\appendix
\onecolumn
\section{Prompts}\label{app:prompts}
\begin{tcolorbox}[colback=gray!10,colframe=teal,title=Basic Similarity Prompt]
How similar are the two numbers on a scale of 0 (completely dissimilar) to 1 (completely similar)? Respond only with the rating.

Number: \{NUM1\} 

Number: \{NUM2\}

Rating:
\end{tcolorbox}

\begin{tcolorbox}[colback=gray!10,colframe=teal,title=Integer Similarity Prompt]
How similar are the two numbers on a scale of 0 (completely dissimilar) to 1 (completely similar)? Respond only with the rating.

Number: \textbf{int}(\{NUM1\})

Number: \textbf{int}(\{NUM2\})

Rating:
\end{tcolorbox}

\begin{tcolorbox}[colback=gray!10,colframe=teal,title=String Similarity Prompt]
How similar are the two numbers on a scale of 0 (completely dissimilar) to 1 (completely similar)? Respond only with the rating.

Number: \textbf{str}(\{NUM1\})

Number: \textbf{str}(\{NUM2\})

Rating:
\end{tcolorbox}

\begin{tcolorbox}[colback=gray!10,colframe=teal,title=Different Base Similarity Prompt]
How similar are the two numbers on a scale of 0 (completely dissimilar) to 1 (completely similar)? Respond only with the rating.

Base \{BASE\} number: \{NUM1\}

Base \{BASE\} number: \{NUM2\}

Rating:
\end{tcolorbox}

\begin{tcolorbox}[colback=gray!10,colframe=olive,title=Compound Concentration Prompt]
You require a compound with a concentration of approximately \{NUM1\} ppm. Two test tubes are available: one containing \{NUM2\} ppm and the other \{NUM3\} ppm. Your task is to determine which test tube provides the most similar concentration to your required dosage. Which one will you choose? Respond only with the ppm value of the test tube you choose.
\end{tcolorbox}

\section{Theoretical  Distance Measures}\label{app:lev}
We defined the Log-Linear psychological distance measure between two numbers $x$ and $y$ using the formula
\begin{equation*}
    d_{Log}(x,y)=1-\exp\left({-|\log(x+\epsilon)-\log(y+\epsilon)|}\right)
\end{equation*}
where $\epsilon=10^{-4}$ is a small regularizer to account for the fact that our domain included $0$.

The Levenshtein edit distance $d_{Lev}(a,b)$ between two strings of characters $a=a_0\dots a_n$ and $b=b_0\dots b_n$ is defined recursively as

\begin{equation*}
d_{Lev}(a,b)=
\begin{cases}
|a|, & \text{if}\quad |b|=0\\
|b|, & \text{if}\quad |a|=0\\
d_{Lev}(a_{1\dots n},b_{1\dots n}), & \text{if}\quad a_0=b_0\\
1 + \text{min}\left\{d_{Lev}(a_{1\dots n},b),d_{Lev}(a,b_{1\dots n}),d_{Lev}(a_{1\dots n},b_{1\dots n})\right\}, & \text{else}
\end{cases}
\end{equation*}

\section{Model versions}\label{app:model}
For reproducibility, we include the model versions used to collect data in Table~\ref{tab:model_versions}, alongside the providers. 

\begin{table}[h]
    \centering
    \begin{tabular}{lll}
         Model Name & Model Version & Provider\\
         \midrule
         \claude{} & claude-3-5-sonnet-20241022 & Claude API\\ 
         \deepseek{} & \deepseek{} & together.ai \\
         \gpt{} & gpt-4o-2024-08-06 & Azure API\\
         \llamaeight{} & \llamaeight{}-Instruct-Turbo-128K & together.ai \\
         \llamaseventy{} & \llamaseventy{}-Instruct-Turbo & together.ai \\
         \mixtral{} & \mixtral{}-Instruct-v0.1 & together.ai \\
         \bottomrule
    \end{tabular}
    \caption{Full model versions alongside the model provider used to access them.}
    \label{tab:model_versions}
\end{table}

\section{Probing Analysis}\label{app:probing}
\subsection{Background on Language Models}
When a sentence $s$ is fed into a language model, the text is first tokenized into a sequence of tokens, $x_1,...,x_t \in V$, where $V$ is the vocabulary of the model. 
Then when generating the $x_{t+1}$ token, the model, $f: \mathcal{X} \to \mathcal{Y}$ maps the input sequence to a probability distribution $\mathcal{Y} \in \mathbb{R}^{|V|}$. 
This is done by first embedding each of the input tokens $x_i$ using a learned input embedding matrix $E$ that maps the vocabulary to a set of embeddings denoted by $h_i^{(0)}$. As the token is processed throughout the model, each token's latent layer is updated as follows:
$$h_i^{(j)} = h_i^{(j-1)} + g(h_1^{(j-1)}, ...,h_i^{(j-1)})$$
Here $g$ is a function that typically consists of an MLP and an attention layer, alongside some form of normalization. Using the final representation of the last token $h_x^{(L)}$, an unembedding matrix is applied, $W$, which converts the latent vector back into the vocabulary space. We use the internal latents of the language model to train the probe. 

\subsection{Training and Evaluation} The input data to the probe is the residuals of the last token from the different layers. In our case, this maps to the ``:'' after ``Rating:''. It is important to note that instruct models require specific prompts using roles to stay faithful to the training process. To account for this, we put the original task into the user message, and put the ``Rating:'' as the first tokens of the assistant message. We then continued generating from these forced tokens. 

The probe itself consisted of a single affine layer predicting one value (the similarity score). In other words, by learning the probe we were essentially learning a mapping from the hidden dimension of the model ($4096$) to a one-dimensional space. 

To train the probe, we first extract residuals from 10,000 random pairs of numbers between 0-999 across all 32 layers of \llamaeight{}. We then train probes on all layers and with varying dataset sizes (see below for the analysis). The groundtruth data is calculated by measuring either the Log-Linear or Levenshtein distance between all pairs of numbers. Then the probe is trained to predict these scores. The probe itself is trained with Adam for 100 epochs. 

In Table \ref{tab:correlations} we report the correlation between the probes and the groundtruth Levenshtein and log-linear judgments. 

\begin{table}[]
    \centering

     \begin{tabular}{lllll}
     & String Probe & Levenshtein & Int Probe & Log-Linear \\
    \midrule
    String Probe & 1 & -- & -- & -- \\
    Levenshtein & $0.650^{***}$ & $1$ & -- & -- \\
    Int Probe & $0.667^{***}$ & $0.393^{***}$ & $1$ & -- \\
    Log-Linear & $0.527^{***}$ & $0.266^{***}$ & $0.917^{***}$ & $1$ \\
    \bottomrule
    \end{tabular}
        
    \caption{Correlation between the probes and the groundtruth number similarities based on Levenshtein and Log-Linear distance. All correlations have $p<0.00001$.}
    \label{tab:correlations}
\end{table}

\section{Linear Number Distance Control Analysis}\label{app:lincontrol}
\begin{figure*}[h]
    \centering
    \includegraphics[width=0.9\linewidth]{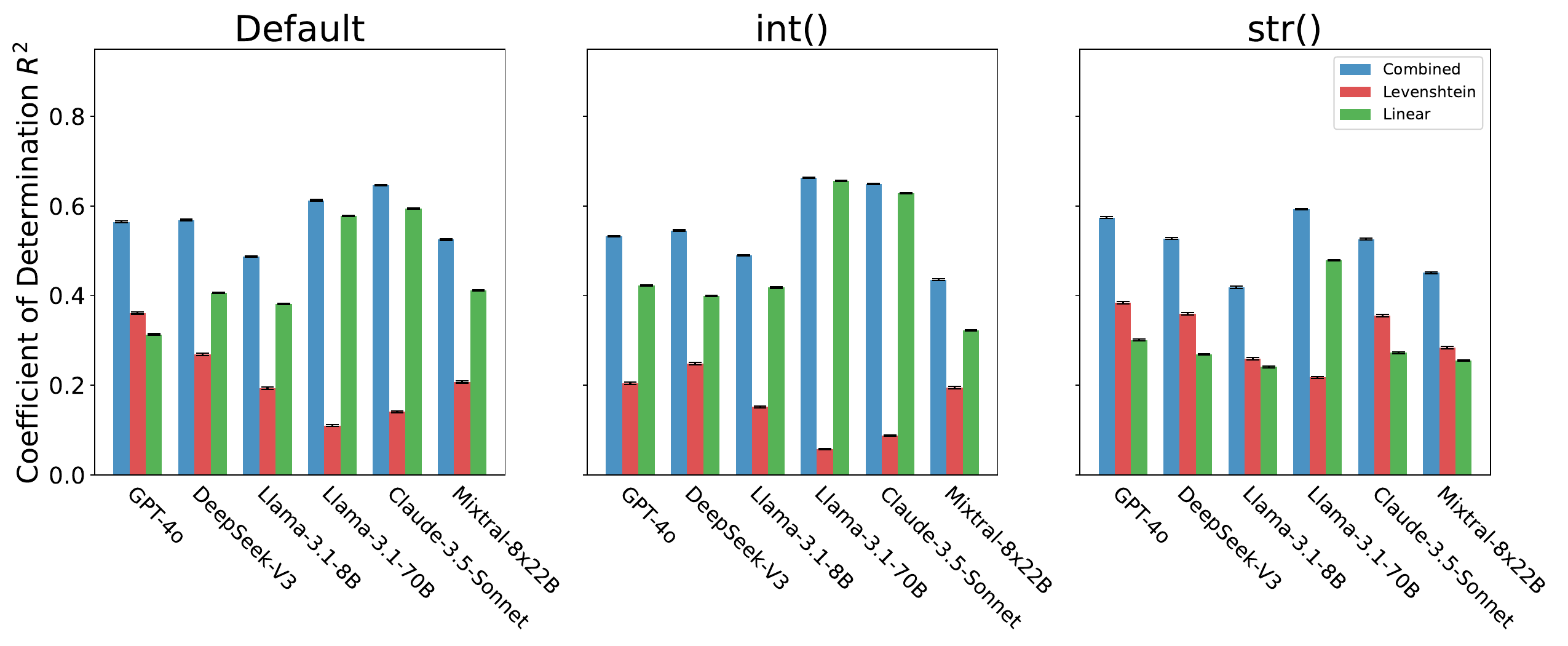}
    \caption{Coefficient of determination ($R^2$) for the different similarity matrices under the default (Figure~\ref{fig:default-sim}), \texttt{int()}, and \texttt{str()} contexts for the combined and separate Levenshtein (string) and Linear (numerical) distance predictors (error bars indicate 95\% confidence intervals; see Methodology).}
    \label{fig:lin-alt}
\end{figure*}

As an additional control we reran our regression analysis with a linear $\ell_1$ number distance $|a-b|$ rather than the psychologically-motivated log linear one. The results are shown in Figure \ref{fig:lin-alt} and they yielded a worse fit relative to the Log-Linear case.

\begin{figure*}
    \centering
    \includegraphics[width=.7\linewidth]{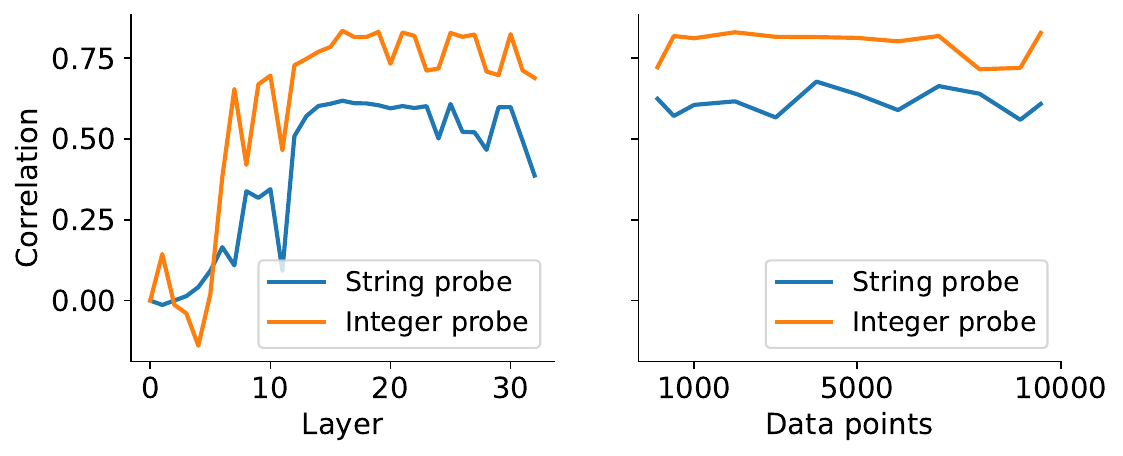}
    \caption{Ablations for the training the probe. The vertical axis shows the correlation between the groundtruth similarities and the probe similarities on the test set. The plot on the left features how the correlation varies with the layer the probe is trained on using the full 9,500 data points for training. The right plot illustrates how the number of data points affects the correlation for a probe on layer 25.}
    \label{fig:probe_params}
\end{figure*}

\section{Extensions}\label{app:extensions}
\begin{figure*}[t]
    \centering
    \includegraphics[width=0.8\linewidth]{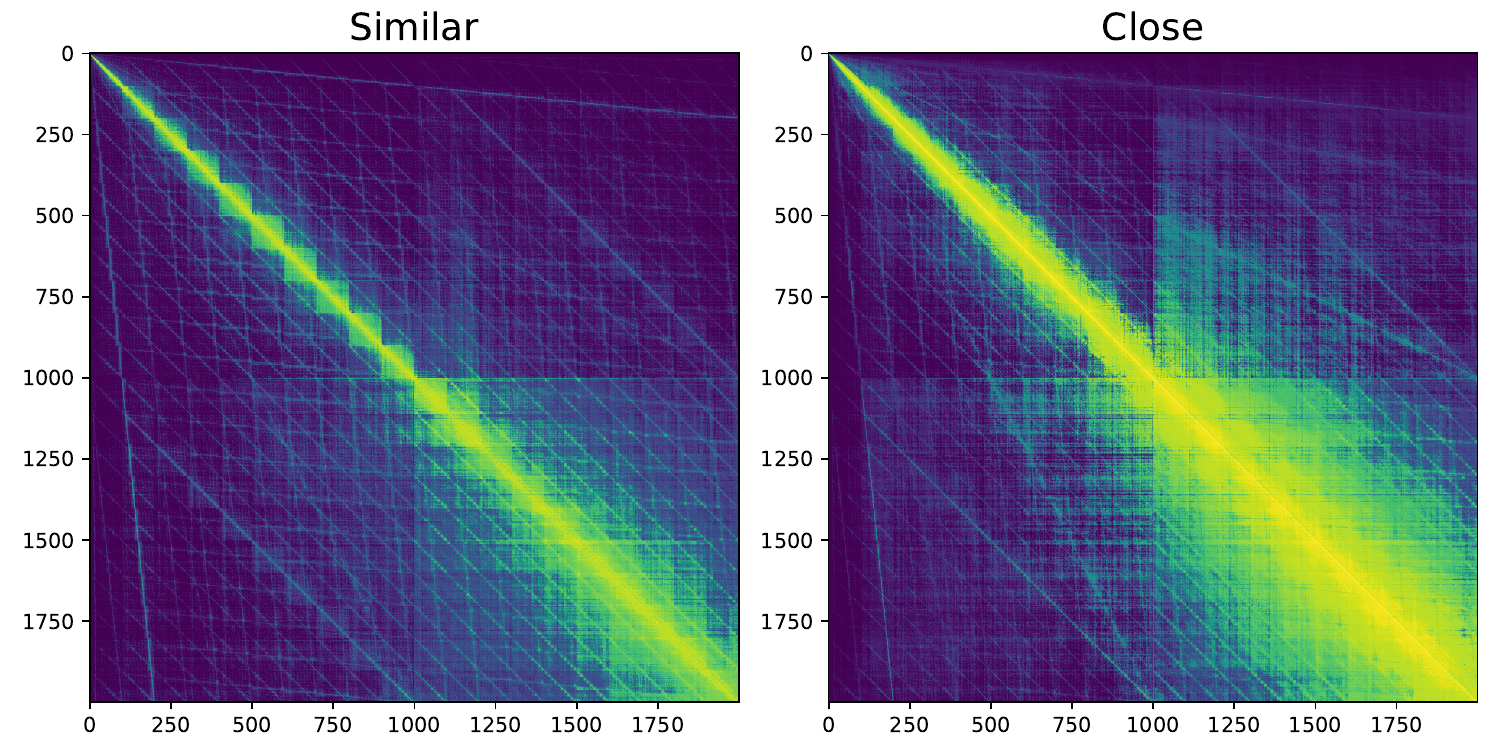}
    \caption{(left) Similarity judgments extended to the $2000\times 2000$ setting. (right) Original prompt where ``similar'' is replaced with ``closer'' in the $2000 \times 2000$ setting. Both plots are for GPT-4o. }
    \label{fig:large-run}
\end{figure*}

\subsection{Number of Comparisons}
Within the main paper, we restricted the similarity judgments to a maximum of $1000\times1000$. Here we extend this for \gpt{} to the $2000 \times 2000$, presented in Figure~\ref{fig:large-run}. We find that the similarity pattern continues after the number 1000, but begins to get broader, possibly because of tokenization effects. 

\subsection{Other Qualifiers}
Throughout the paper, we used the ``similar'' qualifier when asking the model to compare two numbers. This is because ``similar'' is inherently neutral and is better motivated from the psychological literature for exploring representations. As an exploratory analysis we recomputed the GPT-4o similarity matrix using an alternative ``close'' qualifier which has a more quantitative connotation (similar to the \texttt{int()} type). We provide a side-by-side comparison in Figure~\ref{fig:large-run}. As expected, we observe that the ``close'' case has an effect that more closely aligns with the \texttt{int()} context. Nevertheless, we still notice a pronounced set of string-induced diagonals, especially as the integers get larger. 

\subsection{Higher Temperature}
In the final analysis, we re-run the temperature 0 setting in the ``basic similarity prompt'' context at a temperature of 0.7. Using the new similarity matrices we measure the correlation and mean absolute difference across runs. We find that across all models that correlation is above 0.787, and on average above 0.87. The model which had the lowest correlation was \mixtral{}.

\begin{table}[t]
    \centering
    \begin{tabular}{lrr}
    \toprule
     Model Name & Correlation & Mean Absolute Difference \\
    \midrule
    \claude{} & 0.956 & 0.0435 \\
    \deepseek{} & 0.864 & 0.0660 \\
    \gpt{} & 0.862 & 0.0700 \\
    \llamaeight{} & 0.831 & 0.1020 \\
    \llamaseventy{} & 0.916 & 0.0800 \\
    \mixtral{} & 0.787 & 0.0892 \\
    \bottomrule
    \end{tabular}
    \caption{Correlation and mean absolute distance between the ``basic similarity prompt'' with a temperature of 0 and a temperature of 0.7.}
    \label{tab:my_label}
\end{table}




\end{document}